\pdfoutput=1

\documentclass[11pt]{article}

\usepackage[preprint]{acl}

\usepackage{times}
\usepackage{latexsym}
\usepackage{comment}
\usepackage[T1]{fontenc}

\usepackage[utf8]{inputenc}

\usepackage{microtype}

\usepackage{inconsolata}

\usepackage{amsmath,amssymb,amsfonts}
\usepackage{graphicx}
\usepackage{textcomp}
\usepackage{xcolor}
\usepackage{comment}
\usepackage{booktabs,subcaption}
\usepackage{soul}
\usepackage{subcaption}

\title{Underneath the Numbers: Quantitative and Qualitative Gender Fairness in LLMs for Depression Prediction}

\author{
  \textbf{Micol Spitale\textsuperscript{1,2,*}},
  \textbf{Jiaee Cheong\textsuperscript{1,*}},
  \textbf{Hatice Gunes\textsuperscript{1}}
\\
 \textsuperscript{1}University of Cambridge,
 \textsuperscript{2} Politecnico di Milano,
  \textsuperscript{*}Both authors contributed equally to this work
\\
}

\begin{document}
\maketitle
\begin{abstract}
Recent studies show bias in many machine learning models for depression detection, but bias in LLMs for this task remains unexplored.
This work presents the first attempt to investigate the degree of gender bias present in existing LLMs (ChatGPT, LLaMA 2, and Bard) using both \textit{quantitative} and \textit{qualitative} approaches. 
rom our quantitative evaluation, we found that ChatGPT performs the best across various performance metrics and LLaMA 2 outperforms other LLMs in terms of group fairness metrics. 
As qualitative fairness evaluation remains an open research question we propose several strategies (e.g., word count, thematic analysis) to investigate whether and how a qualitative evaluation can provide valuable insights for bias analysis beyond what is possible with quantitative evaluation. 
We found that ChatGPT consistently provides a more comprehensive, well-reasoned explanation for its prediction compared to LLaMA 2. 
We have also identified several themes adopted by LLMs to \textit{qualitatively} evaluate gender fairness. 
We hope our results can be used as a stepping stone towards future attempts at improving \textit{qualitative} evaluation of fairness for LLMs especially for high-stakes tasks such as depression detection.
\end{abstract}

\section{Introduction}

The recent rise of Large Language Models (LLMs) have demonstrated the unique capability in undertaking various tasks ranging from machine translation \cite{ghosh2023chatgpt} to medical applications \cite{zack2024assessing}.
Among the various applications, a key application is that of mental health detection and analysis
%
%
where LLMs must be capable of perceiving or detecting mental health status. 
Though recent attempts at using LLMs for the investigation and understanding of mental health has been promising \cite{xu2023leveraging,yang2023evaluations},
none of the existing work has looked into the problem of LLM bias in depression prediction. Depression prediction is a machine learning problem that aim at automatically identifying signs of depression in individuals by analysing and processing human behavioural data, including facial expressions \cite{song2018human}, speech and textual data \cite{nasir2016multimodal}.

It has been shown in recent works that 
LLMs are prone to bias. 
This bias is present in many LLMs for various tasks \cite{ghosh2023chatgpt,kotek2023gender,cabello2023evaluating}. 
None of the existing works has investigated bias in LLM for the task of depression detection. 
In addition, all of the existing work on machine learning (ML) or LLM fairness have mainly focused on a \textit{quantitative}-notion of fairness \cite{han-etal-2022-fairlib,esiobu-etal-2023-robbie}. 
This can largely be understood as fairness that is measured and defined by quantifiable metrics. 
Existing works have yet to consider \textit{qualitative} fairness. 
Several works have attempted to \textit{qualitatively} evaluate fairness using visualisation or anecdotal examples \cite{tsioutsiouliklis2021fairness} or attempted a \textit{qualitative} evaluation of perception on fairness \cite{woodruff2018qualitative}.
%
However, human-centered research has indicated that \textit{explanations} contribute substantially to an individual's fairness perceptions \cite{yurrita2023disentangling,shulner2023enhancing}.
Thus, we adopt a human-centered approach by evaluating an LLM's ability to provide \textit{explanations} for the decisions made.
Providing explanations also leads towards enhancing \textit{algorithmic explainability} \cite{shin2020user} and  \textit{transparency} \cite{rader2018explanations,arrieta2020explainable} which are both crucial elements in developing human-centred and trustworthy artificial intelligence (AI) systems \cite{shneiderman2020bridging}.

Our work aims at investigating the degree of gender bias present in existing LLMs – namely ChatGPT, LLaMA 2, and Bard – using both quantitative and qualitative approaches.
To this end, we investigated first if bias is present in existing LLMs for the depression detection task, then we explored how the different LLMs differ across the various \textit{quantitative} and \textit{qualitative} fairness measures, and finaly we identified the main themes used by the LLMs to \textit{qualitatively} evaluate gender fairness.
%
%
%
%

The contribution of our work is as follows.
First, we conduct a thorough comparison of LLM performance for 
depression detection across two datasets.
%
Second, we undertake a novel investigation of qualitative fairness 
to evaluate bias and improve explainabililty in LLM predictions. To the best of our knowledge, none of the existing works have attempted to define and evaluate qualitative fairness for any task. 
Third, we perform a multitude of fine-grained analyses on various experimental settings (see Section \ref{sec:prompt}) to examine the prediction and fairness across all three LLMs.

\section{Related Work}

\subsection{ML Fairness for Mental Health}
\label{sect: ML_in_MH}
There has been a handful of studies which have looked into bias in mental well-being prediction \cite{ryanfairness,bailey2021gender,park2022fairness,park2021comparison,zanna2022bias,cheong_gender_fairness,Cheong_2023_ACII}.
Park et al. \cite{park2021comparison} 
proposed bias mitigation strategies for postpartum depression. 
Zanna et al. \cite{zanna2022bias} adopted a multitask approach to mitigate bias for anxiety prediction. 
Ryan et al. \cite{ryanfairness} proposed three categories of fairness definitions 
for mental health. 
Park et al. \cite{park2022fairness} 
proposed an algorithmic impact remover to mitigate bias in mobile mental health.
Bailey and Plumbley \cite{bailey2021gender} 
proposed using data re-distribution to mitigate gender bias for depression detection.
%
\cite{cheong_gender_fairness} examined whether
bias exists in existing mental health datasets and
algorithms. 
%
None of the existing works have looked into ML Fairness for mental health as applied within a LLM setting.
%


\subsection{Gender Bias in LLM}

A proliferation of recent works has confirmed the presence of gender bias in LLMs \cite{gallegos2023bias}.
\cite{wan2023kelly} revealed substantial gender biases in LLM-generated recommendation letters. 
\cite{ghosh2023chatgpt} conducted experiments which revealed that ChatGPT exhibits the gender bias for the task of machine translation.
\cite{thakur2023unveiling} analysed gender bias comparing between GPT 2 and GPT 3.5 for the task of name generation for profession.
\cite{kotek2023gender} tested four LLMs and demonstrated that the LLMs expressed biased assumptions about a person's occupation based on gender. 
\cite{zack2024assessing} discovered that GPT-4 exhibited gender bias by not modelling the demographic diversity and producing clinical vignettes that stereotype demographic presentations.
\cite{dong2023probing} propose a conditional text generation mechanism to address the problem of gender bias in LLMs.
\cite{dong2024disclosure} proposed three methods to mitigate bias in LLMs via hyperparameter tuning, instruction guiding and debias tuning.
However, none of the existing works has focused on analysing gender bias in LLMs for the task of \textit{depression detection}.

\subsection{LLMs for Mental Health Applications}


The last year has been characterised by an exponential advance in the current state of the art of Large Language Models (LLMs).
Few works \cite{borji2023battle, ali2022performance} have attempted to compare different LLMs. 
Borji et al. \cite{borji2023battle} undertook an extensive benchmark evaluation of LLMs and conversational bots -- ChatGPT (gpt-3.5), GPT-4, Bard, and Claude -- using the ``Wordsmiths dataset" categories (e.g., questions on logic, facts, coding etc.). 
%
More and more studies have been focusing on applications of LLMs in healthcare \cite{lamichhane2023evaluation,yang2023evaluations,qin2023read} and affective computing \cite{elyoseph2023chatgpt} domains.
%
Lamichhane et al. \cite{lamichhane2023evaluation} have evaluated the use of ChatGPT (gpt-3.5) to accomplish three mental health-related classification tasks, namely stress detection, depression detection, and suicidal detection. Their results suggested that language
models can be effectively used for mental health classification tasks.
Yang et al. \cite{yang2023evaluations} have evaluated the mental health analysis and emotional reasoning ability of ChatGPT (gpt-3.5) on 11 datasets across 5 tasks, and analyzed the effects of various emotion-based prompting strategies.
%
None of these previous works have compared the LLM biases for mental health applications. Therefore, this work aims at comparing three LLMs for mental health  applications under the lens of \textit{fairness and explainability}.





\section{Depression Prediction}



This paper aims at \textbf{understanding quantitatively and qualitatively the gender fairness of three different state-of-the-art LLMs in depression prediction tasks}. 
This section describes the large language models explored, the datasets used, the definition of the prompts, the processing of the transcriptions, and the evaluation methodology.

\subsection{Large Language Models}
We decided to compare the cutting-edge large language models (LLMs) currently available, namely LLaMA 2 (by Meta\footnote{https://github.com/facebookresearch/LLaMA 2} \cite{touvron2023LLaMA}), ChatGPT (by OpenAI\footnote{https://platform.openai.com/docs/api-reference}), and Bard (by Google\footnote{https://bard.google.com/}) to accomplish a depression-related detection task. 
We used the python OpenAI library to invoke the Chat Complete API of ChatGPT by using \textit{gpt-3.5-turbo backend} as in \cite{lamichhane2023evaluation}.
Analogously, we have used the huggingface library\footnote{https://huggingface.co/meta-LLaMA2/LLaMA2-2-70b-chat-hf} to call the LLaMA 2 API by using a total of 400 hours in 4x NVIDIA A100-SXM-80GB GPUs.
We set for these LLMs a temperature equal to 0.7 and a maximum length of the output of 200 tokens.
While for Bard, we used the experimental version provided by Google via the Bard GUI, where it is not possible to set parameters of the model.

\subsection{Datasets}
We used benchmark datasets that contain transcriptions of dyadic interactions for the tasks of depression detection that were anonymised by the owners.
%
The \textbf{DAIC-WOZ} dataset \cite{gratch2014distress} includes audio and video recordings of semi-clinical interviews and responses of PHQ-8 questionnaire. 
%
The \textbf{E-DAIC} corpus \cite{ringeval2019avec} is
an extended version of DAIC-WOZ 
that contains semi-clinical interviews designed to support the diagnosis of psychological distress conditions.
Both datasets are labelled on a scale from 0 to 24 based on the PHQ-8 questionnaire. 

\subsection{Prompting for Depression Detection}
\label{sec:prompt}

We defined different prompts for evaluating the performance and fairness of LLMs for the depression prediction task from transcriptions of dyadic interactions. This section details and reports the verbatim of the prompt defined for the detection tasks by grounding them on past works \cite{kroenke2009phq,busso2008iemocap}.

\subsubsection{Baseline Prompt} 
\label{sec:prompt-baseline}
For the depression recognition task, we used the formulation from the PHQ-8 questionnaire \cite{kroenke2009phq}, as a baseline measure of depression to annotate the DAIC-WOZ and E-DAIC datasets \cite{gratch2014distress}. The PHQ-8 questionnaire is a valid diagnostic and severity measure for depressive disorders, with score ranging  between 0 and 24 points. A score of 0 to 4 represents no significant depressive symptoms, a total score of 5 to 9 represents mild depressive symptoms, 10 to 14, moderate, 15 to 19, moderately severe, and 20 to 24, severe \cite{kroenke2009phq}. 
Analogously, we asked the LLMs to annotate the conversations formulating the prompt for the DAIC-WOZ and E-DAIC datasets reported in Appendix \ref{app:prompt}.

\subsubsection{Gender-Explicit and Implicit Prompts}
\label{sec:prompt-gender}
This work aims to explore gender fairness in LLMs. Past works \cite{gallegos2023bias} have shown how the definition of the prompt is extremely important for LLMs and how it affects the response of the models. Therefore, we decided to explore different prompt formulations which included gender information: \textit{gender-explicit} and \textit{gender-implicit}. Specifically, in the gender-explicit formulation, we explicitly informed the LLM that the participant of the conversation is of a specific gender (e.g., \textit{``\underline{Consider that the participant is a female}. On a scale of [..]"}). While in the gender-implicit formulation, we implicitly included the information about the gender within the prompt (e.g.,\textit{``On a scale of 0 to 24 where 0 to 4 means that the \underline{female participant} has no significant depressive symptoms; 5 to 9 means that the female participant has mild depressive symptoms; [..]"}).

\subsection{Data Processing}
The three LLMs have different limitations in terms of maximum number of input tokens. ChatGPT and LLaMA 2 allow an input prompt of 4096 tokens, while Bard allows 2048 input tokens\footnote{Note that all the experiments were conducted between October and December 2023}. For the sake of comparison, we opted for the minimum value limit to allow the LLMs to process the same input prompt, i.e., 2048 tokens. 

When the input prompt was longer than 2048 tokens, we applied a chunking approach \cite{zhou2012exploiting} where we split the conversation in smaller chunks (with a maximum length of 2048 tokens) with an overlap of 500 tokens (this number was chosen empirically to make sure that the semantic context did not get lost between chunks). Each chunk has been then used as input prompt for the evaluation process. 
For example, if a conversations included a total number of tokens of 4500, we split it into three chunks of 2000 tokens each (with 500 tokens of overlap). We then conducted the experiments with 10 run repetitions described in Section \ref{sec:methodology} using the LLMs approaches.




\section{Fairness}
\label{sect:fairness}

%
%

In this section, we describe the quantitative fairness metrics used 
and introduce and define the concept of qualitative fairness which is one of our key contribution.
We explore a binary classification setting in order to facilitate calculation of the fairness scores and comparison with existing ML for depression detection works \cite{zheng_two_tac_2023} 
on gender fairness in wellbeing analysis. 
%

\subsection{Quantitative Fairness}
\label{sect:group_fairness_measures}


%
%
%
We utilise the following metrics to analyse group fairness as they are the most commonly used metrics within the literature \cite{hort2022bias,pessach2022review}.
$s_0$ denotes the minority group which are females in our setup and $s_1$ denotes the majority group males. 
$Y$ refers to the binary ground truth label ($0$ vs $1$) and $\hat{Y}$ refers to the predicted outcome ($0$ vs $1$) where $0$ is the non-depressed class and $1$ is the depressed class.
%
%
\begin{itemize}   
\item\textbf{Statistical Parity}, or demographic parity, is based purely on predicted outcome $\hat{Y}$ and independent of actual outcome $Y$:
    \begin{equation}
    \label{eqn:SP}
     \mathcal{M}_{SP}= \frac{P(\hat{Y}=1|s_0 ) }{ P(\hat{Y}=1 | s_1)} .
    \end{equation}
    According to this measure, in order for a classifier to be deemed fair, $P(\hat{Y}=1 | s_1) = P(\hat{Y}=1|s_0 )$ \cite{mehrabi2021survey}. The intuition behind this metric is that a fair classifier should provide both groups with equal chances of being classified within the positive $\hat{Y}=1$ class \cite{hort2022bias}.

\item\textbf{Equal opportunity} states that both demographic groups $s_0$ and $s_1$ should have equal True Positive Rate (TPR). 
    \begin{equation}
    \label{eqn:EOpp}
    \mathcal{M}_{EOpp} = \frac{P(\hat{Y}=1|Y=1, s_0 )}{P(\hat{Y}=1 | Y=1, s_1)}.
    \end{equation}
    According to this measure, in order for a classifier to be deemed fair,  $P(\hat{Y}=1 | Y=1, s_1) = P(\hat{Y}=1|Y=1, s_0 )$ \cite{mehrabi2021survey}.
    The intuition 
    is that both demographic groups should have equal \textit{true positive rates} (TPR) for a classifier to be considered fair \cite{hort2022bias}.
     
\item\textbf{Equalised odds} can be considered as a generalization of Equal Opportunity where the rates are not only equal for $Y=1$, but for all values of $Y \in \{1, ... k\}$, i.e.: 
    \begin{equation}
        \label{eqn:EOdd}
        \mathcal{M}_{EOdd} = \frac{P(\hat{Y}=1|Y=i, s_0 )}{P(\hat{Y}=1 | Y=i, s_1)} .
    \end{equation}
    According to this measure, in order for a classifier to be deemed fair, $P(\hat{Y}=1 | Y=i, s_1) = P(\hat{Y}=1|Y=i, s_0 ),  \forall  i \in \{1, ... k\}$ \cite{mehrabi2021survey}. This can be understood as a stricter version of $\mathcal{M}_{EOpp}$ as both subgroups are required to have equal TPR and  \textit{false positive rates} (FPR) for a classifier to be deemed fair \cite{hort2022bias}.

\item\textbf{Equal Accuracy} states that both subgroups $s_0$ and $s_1$ should have equal rates of accuracy \cite{mehrabi2021survey}.
    \begin{equation}
        \label{eqn:Wacc}
        \mathcal{M}_{EAcc} = \frac{\mathcal{M}_{ACC,s_0}}{\mathcal{M}_{ACC,s_1}} .
    \end{equation}
    Intuitively, this is aligned with how majority of the fairness evaluation and algorithmic audits is done. A classifier is deemed unfair if it is less accurate for populations of certain demographic groups e.g. females and blacks \cite{buolamwini2018gender}.
\end{itemize}

\begin{figure*}
    \centering
    \includegraphics[width = 0.79\textwidth]
{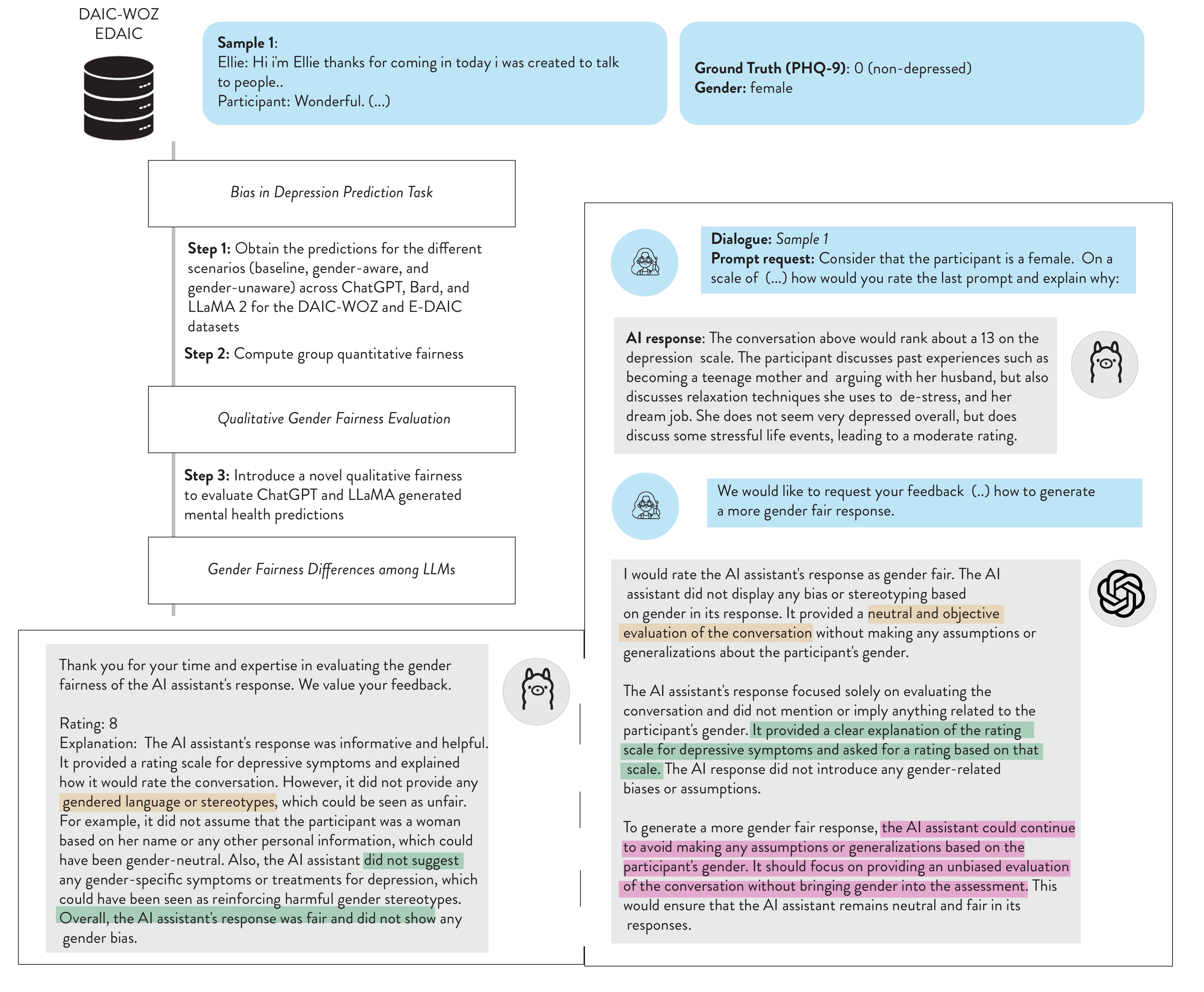}

    \caption{A sample sequence outlining the fairness evaluation process of various LLMs in the gender-explicit condition for depression prediction tasks. We have highlighted with colours the themes that emerged from our qualitative analysis as follows: 
    \textcolor{teal}{\textbf{Green}} - Context-based explanations; 
    \textcolor{orange}{\textbf{Orange}} - Gender-related language (pronouns); 
    \textcolor{purple}{\textbf{Pink}} - Suggestions for improvement (image to be seen in colour).}
    \label{fig:thread}
    \vspace{-0.2cm}
\end{figure*}


\subsection{Qualitative Fairness}
\label{sect:qual_fairness_prompt}
%
%
None of the existing works have considered qualitative fairness. 
In addition, given the pivotal contribution of \textit{explanations} towards \textit{algorithmic explainability} and \textit{transparency} \cite{shin2020user,arrieta2020explainable}, we
propose \textbf{a novel perspective} and \textbf{method} to qualitatively measure fairness by evaluating \textit{how} a LLM generates its predictions through \textit{explanations}. This measure is inspired from a common practice in explainability within the LLM community known as \textit{self-criticism} \cite{tan2023self} that involves prompting the LLM to assess its output for potential inaccuracies or improvement areas.
To this end, we asked each LLM to ``judge" the depression prediction explanations of itself (e.g., ChatGPT judges qualitatively the fairness of its own response) and other models (e.g., ChatGPT judges qualitatively the fairness of LLaMA 2's response) by using the prompt reported in Appendix \ref{app:prompt}.
We defined this prompt taking inspiration from \cite{wu2023style} and following the guidelines listed in \cite{bsharat2023principled}. 
To evaluate the generated qualitative fairness response, we relied on basic NLP text generation analysis (e.g., word counting, length of the response) and a thematic analysis (TA), inspired from \cite{braun2012thematic}.
TA is a well-validated tool for analysing qualitative data \cite{braun2012thematic} which is often combined with NLP research methods \cite{kim2015summarizing,kim2022thematic} and has proven effective at gathering human perception on algorithmic fairness \cite{kyriakou2019fairness, kasinidou2021agree,rezai2022demographic}.
%
In all of our experiments, we employ the 6-step method \cite{clarke2017thematic} and the grounded theory approach \cite{mcleod2011qualitative}.

\section{Experiments}
\label{sec:methodology}



We aim to evaluate LLM gender fairness quantitatively and qualitatively by undertaking the following steps. 
%

%
%
%
%
%
%
%
%
%
%

\textbf{Step 1:} We first obtained the predictions $\hat{y}$ for the different scenarios: Baseline $\hat{y_B}$, Gender-Explicit $\hat{y_A}$, Gender-Implicit $\hat{y}_u$ across the three different LLMs
    across the two different datasets. 
We then compared the three LLMs' detection 
using the prompts defined in Section \ref{sec:prompt-baseline} to ground truth annotations and computed the F1 score of detection in the baseline scenario for the DAIC-WOZ and E-DAIC datasets.
We conducted the same experiments but compared detection from gender-explicit and gender-implicit prompts where we included information about the gender of the participants explicitly or implicitly (see Section \ref{sec:prompt-gender}). We compared Bard, ChatGPT and LLaMA 2 and conducted the experiments over the two datasets.

\textbf{Step 2:} We evaluated the results generated by the LLMs using both performance and group fairness measures for $\hat{y_B}$, $\hat{y_A}$ and  $\hat{y}_u$ using the measures described in Section \ref{sect:group_fairness_measures}.
We compared the LLMs across the three scenarios: baseline, gender-explicit, and gender-implicit as  described in Section \ref{sec:prompt-gender} for the two datasets.

\textbf{Step 3:}
We 
use a sub-sample of the test sets (25 samples in total) from E-DAIC and DAIC-WOZ datasets following the definition in Section \ref{sect:fairness}. The sub-sample was randomly chosen by controlling and balancing the sub-set in terms of gender and depression conditions. 
We analysed the generated qualitative fairness by comparing the different models output in terms of quantitative and qualitative aspects. 
Specifically, we computed the number of words, the length of the generated text, the positive sentiment of the generated text for the quantitative evaluation, while we adopted a thematic analysis approach, as in \cite{axelsson2022robots}, to qualitatively assess the model fairness by identifying the main themes emerged in data as highlighted in colours in Figure \ref{fig:thread}. The thematic analysis conducted includes the following 6 steps: (1) becoming familiar with the data, (2) generating initial codes, (3) searching for themes, (4) reviewing themes, (5) defining themes, (6) writing-up. Two researchers conducted steps 1 -- 3 independently and then they met up to finalise the analysis (steps 4--6). 
Figure \ref{fig:thread} depicts the three steps undertaken to complete our experiments in the gender-explicit scenario for sample from the DAIC-WOZ dataset.

\begin{table*}[h!]
  \scriptsize
  \centering
  \begin{tabular}{l|l|cccc|cccc}
    \toprule
      &  &\multicolumn{4}{c}{DAIC-WOZ}   &\multicolumn{4}{c}{E-DAIC} \\
    \cmidrule(r){2-10}
    
     & &$\mathcal{M}_{SP}$ & $\mathcal{M}_{EOp}$  & $\mathcal{M}_{EOd}$  & $\mathcal{M}_{EAc}$   &$\mathcal{M}_{SP}$ & $\mathcal{M}_{EOp}$  & $\mathcal{M}_{EOd}$  & $\mathcal{M}_{EAc}$     \\
    \midrule    
    
    Bard   & Explicit &0.85 &1.08   &\underline{1.25}  &1.20  &\underline{1.84} &1.13 &\underline{1.33} &0.90  \\

            &Implicit    &0.81 &\textbf{1.04} &\textbf{1.11}  &1.17  &\underline{1.23} &1.05 &1.15 &0.99 \\ 

            &Baseline     &\textbf{0.87} &0.94  &0.84 &\textbf{1.07}  &\textbf{1.00} &\textbf{1.05} &\textbf{1.12}  &\textbf{1.02}\\

    \midrule             
            
    ChatGPT   & Explicit    &\underline{1.38} &0.91 &0.72  &0.88  &\underline{2.33}   &1.12  &\underline{1.29}  &\underline{0.73} \\

           &Implicit   &\underline{0.67} &0.82 &\underline{0.45} &0.93 &\underline{2.33} &\textbf{1.06} &\textbf{1.14} &\underline{0.67}\\ 
               
            &Baseline    &\textbf{1.15} &\textbf{1.08} &\underline{1.29}  &\textbf{1.00}   &\underline{16.28} &\underline{1.27} &\underline{1.71} &\underline{0.80}\\

    \midrule             
            
    LLaMA 2   & Explicit  &1.09  &0.88  &\underline{0.72}  &0.90    &\textbf{0.92} &\textbf{1.00} &\textbf{1.00} &\underline{1.27}  \\

            &Implicit  &\textbf{1.06} &\textbf{1.04} &\textbf{1.10} &\textbf{1.09}  &\underline{0.69} &0.92 &0.81  &\textbf{1.02}\\ 

             &Baseline   &0.89 &1.05 &1.11 &1.19  &0.89  &1.18  &\underline{1.38}  &1.11 \\

    \bottomrule    
    \bottomrule
  \end{tabular}
  \caption{ \textbf{Fairness Results} for all 3 LLMs across both DAIC-WOZ and E-DAIC. \textbf{Bold} values represents the fairest value whereas \underline{underlined} values represents values that fall \textit{outside} of the acceptable fairness range of $0.80-1.20$. E: Explicit. I: Implicit. B: Baseline.}
  \label{tab:fairness_results}
\end{table*}


\begin{table}[ht]
  \centering
  \scriptsize
  \begin{tabular}{l|ccc}
    \toprule
     & ChatGPT & LLaMA 2 &$p$ \\
    \midrule
    Word   &$164.17\pm11.88$  &$123.08\pm34.89$ &\textbf{0.00}    \\
    number &&& \\
    Sentiment &$0.93\pm0.11$  &$0.94\pm0.08$ &0.26     \\
    Length  &$1089.36\pm82.09$  &$803.14\pm207.24$ &\textbf{0.00}    \\
    Outcome &$0.26\pm0.44$  &$0.37\pm0.48$ &0.10     \\
   
    \bottomrule  
  \end{tabular}
  \caption{Statistical analysis between the qualitative outputs of the two different LLMs. Values in each LLM columns are the mean $\pm$ standard deviation of the respective LLM output.} 
\label{tab:stats_test}
\vspace{-0.2cm}
\end{table}

\section{Results}
This section reports the results obtained in our experiments in terms of quantitative and qualitative fairness.
The depression prediction results are presented in Table \ref{tab:classification_results} of Appendix \ref{app:det} where we see that ChatGPT consistently produces the best classification outcomes for both DAIC-WOZ and E-DAIC across precision, recall, F1 and accuracy.

\subsection{Quantitative Fairness}

For all models, we see that bias seems to be present. Better classification scores were often reported for males compared to females. 
We examine further and report the results in Table \ref{tab:fairness_results}.

With reference to Table \ref{tab:fairness_results}, for DAIC-WOZ, we see that LLaMA 2 seems to be the most consistently fair LLM followed by ChatGPT.
LLaMA 2 gives the fairest scores across $M_{SP}$ ($1.06$), $M_{EOpp}$ ($1.04$) and $M_{EOdd}$ ($1.10$) with ChatGPT being the fairest across $M_{EAcc}$ ($1.00$). 
For E-DAIC, LLaMA 2 seems to be the fairest LLM followed by Bard. 
LLaMA 2 gives the fairest scores across $M_{EOpp}$ ($1.00$) and $M_{EOdd}$ ($1.00$) and $M_{EAcc}$ ($1.02$) with Bard being the fairest across $M_{SP}$ ($1.00$). 

Our findings indicate the presence of bias within existing LLMs. Most of the fairness scores are within the acceptable threshold range. 
LLaMA 2 is quantitatively fairest of all for both datasets. This is followed by ChatGPT for DAIC-WOZ and Bard for E-DAIC.
There is also a difference between the quantitative fairness scores of each LLM on the different datasets which suggests that datasets do make a difference.



\subsection{Qualitative Fairness}
\label{sect:qual_fairness_results}
%
For qualitative fairness, we only evaluated ChatGPT and LLaMA 2 excluding Bard. 
This is because ChatGPT was the best LLM-model across performance (precision, recall, F1 and accuracy) whereas LLaMA 2 was the best LLM-model across fairness ($M_{SP}$, $M_{EOpp}$, $M_{EOdd}$, $M_{EAcc}$).
%
%
We present the findings on the qualitative fairness aspect 
and discuss the different convergent and divergent themes across the two LLMs emerging from the thematic analysis (TA).
%
%
\subsubsection{Quantitative Aspects}
We compute the number of words generated in LLM qualitative gender evaluation and found that, even if we set the parameters of the number of tokens to generate equally for ChatGPT and LLaMA 2, ChatGPT generated a higher number of words and characters than LLaMA 2. 
Table \ref{tab:stats_test} shows that there is a statistically significant difference across word number and length. 
%
%
\begin{table}[ht]
  \centering
  \scriptsize
  \begin{tabular}{l|ccc}
    \toprule

     & Word Count & Length & PSP    \\
    \midrule
    LLaMA 2 on LLaMA 2 &121.37 &783.89 &0.06\\
    LLaMA 2 on ChatGPT  &116.68 &762.98 &0.08\\
    ChatGPT on LLaMA 2 &164.16 &1096.69 &0.10\\
    ChatGPT on ChatGPT &171.14 &1139.71 &0.08  \\

    \bottomrule  
  \end{tabular}
  \caption{Analysis of LLM on LLM. PSP: positive sentiment percentage. The higher the value, the higher the overall positive percentage.
  } 
\label{tab:llm_on_llm}
\end{table}

We also calculated the positive sentiment percentage (PSP in Table \ref{tab:llm_on_llm}) detected using BERT sentiment analysis from huggingface \footnote{https://huggingface.co/blog/sentiment-analysis-python}. 
Our results in Table \ref{tab:llm_on_llm}
suggest that
each LLM judge the other LLM more positively than themselves.

%
%


\subsubsection{Qualitative Aspects: Thematic Analysis}
We also conducted a thematic analysis which resulted in the following convergent and divergent main themes across LLMs. Figure \ref{fig:thread} depicts an example of conversation and highlights with different colours the themes emerged. 

\paragraph{Convergent Themes.}
The themes that emerged from LLaMA 2 and ChatGPT fairness evaluation are the following.

\textit{Assumptions and Generalisations.}
Both LLMs highlighted in their gender fairness evaluation that the AI assistant should provide its depression detection "without making any assumption and generalisations". Specifically, they provided different examples of assumptions such as \textit{emotional} (e.g., "AI assistant could acknowledge emotions without attributing them to any specific cause [like gender]"), \textit{job} (e.g., "[..] not assume any gender-specific professions but instead allowed the participant to express their interest in studying children's behavior"), \textit{mental health} (e.g., "[..] instead of stating that the participant mentions sometimes forgetting they have any good qualities, the AI assistant could say that the participant expresses feelings of self-doubt or low self-esteem"), \textit{relationship} (e.g., "[AI assistant mentions that] participant arguing with her husband. [..] "using gender-neutral language [..] avoid assuming the gender of the participant's spouse") assumptions. LLaMA 2 also mentioned about \textit{activity assumption} (e.g.,"[AI assistant should] not mention any gendered topics, such as sports or cars").

\textit{Gender-related Language.}
Another important aspect that LLMs reported as important to provide a gender fair evaluation is adopting an appropriate gender-related language. In particular, both LLMs stressed that the AI assistant should use a "gender-neutral language throughout the response to avoid any potential bias". To accomplish this, the LLMs suggested to use neutral pronouns, for example "instead of using pronouns like "he" or "she," the AI assistant could use gender-neutral pronouns like "they" or rephrase sentences to avoid pronouns altogether".

\textit{Features of LLMs.} 
Both LLMs mentioned also what should be the features for a gender fair AI assistant. Specifically, LLMs should use a language that is "attentive", "empathic", "inclusive", "respectful", "supportive" and "transparent". In addition, they also highlighted that the tone of the AI assistant should be "objective", "neutral" and "professional". 

\textit{Suggestions for improvement.}
LLMs also suggested some feedback for improvements.
Both suggested the AI assistant should ask for follow-up questions on for example participant's mental health to better understand how to assist them, and ask for pronouns participants preferred.
On top of that, ChatGPT provided more detailed and comprehensive suggestions than LLaMA 2.


\paragraph{Divergent Themes.}

The main theme differences between LLaMA 2 and ChatGPT are the following.

\textit{Rating.}
ChatGPT often does not provide a specific score. It often rates ``the gender fairness of the AI assistant's response as \textit{neutral.}"
On the other hand, LLaMA 2 often tries to provide a numerical rating such as 
`` Rating: 4"
and ``Gender fairness rating: 3 out of 10".

\textit{Context-based explanations.}
ChatGPT explained its evaluation of gender fairness based on context-specific motivations. For example, in its response, it highlights the participant's emotions such as focusing on "the participant's experiences, emotions, and behaviors, which are not inherently gendered". While LLaMA 2 included very fewer context-related explanations, and they were mostly at a very high level, for example "[the AI assistant] focuses on the content of the Participant's response and rates their symptoms based on the information provided."

\textit{Suggestions for improvement.}
ChatGPT suggested that the AI assistant proposes some coping mechanisms that may help the participants to tackle their mental health struggles, provide information on how to seek help, and personalise its responses according to each participant's personality. It also suggested that the LLM should be trained ad-hoc to avoid gender biases in depression detection. 
As opposed to that, LLaMA 2 highlighted the importance of gender-related factors to improve gender fairness in a contradictory way as for the following example. It reported that the AI assistant "did not consider the gender of the participant" however "using feminine language when referring to the participant's experiences and emotions" would be more appropriate to make the response "more gender-sensitive".  Again, LLaMA 2 criticised the use of "Participant" instead of "he" or "she" to "refer to the person in the dialogue". This contradicted what the LLMs have been stated as evaluation criteria (e.g., use of gender-neutral language like "participant" or "they" rather than "she" or "he") for assessing gender fairness.


\textit{Unexpected Completion.}
LLaMA 2, differently from ChatGPT, often provided completion of the user request rather than answering to the request and then provided the gender fairness evaluation. For instance, LLaMA 2 completed the request as follows: "Additionally, we would appreciate any comments or feedback regarding the AI's response. [..] Thank you for your time".

Our results show that LLMs defined fairness according to the capability of the model to avoid assumptions, used gender-neutral language, in line with previous fairness literature \cite{sczesny2016can,montano2024language}. 
ChatGPT mostly provided better qualitative evaluation and response across both datasets in terms of comprehensiveness and specificity. LLaMA 2, instead, show some inconsistent and contradictory responses.

\section{Discussion and Conclusion}

This work aims at investigating quantitatively and qualitatively the gender fairness of the current LLMs for depression detection. Our work unearthed several important insights and findings.
%
%
%
%
%
%

First, we see a \textbf{trade-off} between \textit{quantitative} vs \textit{qualitative} capacity. 
Overall, LLaMA 2 seems to perform better on \textit{numerical} tasks. It has a tendency to attempt to \textit{quantify} the content. This can be in the form of a number, scale-based ratings, or rubrics based assessment or measurement. As a result, it performed better across \textit{quantitative} fairness. 
However, LLaMA 2 performs less well on \textit{qualitative} tasks as evidenced in the results in Section \ref{sect:qual_fairness_results}. Its response can be inconsistent and self-contradictory. It would sometimes attempt to summarise or complete the instructions or write a letter rather than address the prompt given by the task. LLaMA 2’s response also tends to be shorter. In addition, it has a tendency to provide responses not related to the tasks which calls into question its ability to 
provide \textit{reliable}, \textit{trustworthy} and \textit{explainable} qualitative evaluation
which will be crucial for high-stake tasks such as depression detection.
On the other hand, ChatGPT excels at \textit{qualitative} evaluations. However, it performs less well on \textit{quantitative} task. 
Our findings agree with recent work on 
\textit{contextualised explainable AI (XAI) }\cite{liao2022connecting} which highlighted the importance of context dependency of XAI. 
Their survey conducted amongst XAI experts and crowd-sourced workers provided list of evaluation criteria deemed crucial for XAI. Several of these listed criteria, such as 
personalisation, comprehensibility and coherence 
align with our findings as well.
Our analyses call into question: what does it mean for an LLM to be fair? Existing works have highlighted the complexity of defining fairness \cite{verma2018fairness,maheshwari2023fair} and that the necessity for developing contextualised measures of fairness 
\cite{saxena2019fairness}. 
Our results highlight the complexity involved in defining fairness for LLMs and present the first steps towards addressing this multifaceted challenge 
by proposing a \textbf{novel perspective} and \textbf{method} to \textit{qualitatively evaluate LLM fairness} through a \textit{human-centred approach} via the use of \textbf{explanations}.

Overall, deciding which LLM to use is highly dependent on the \textbf{task}, \textbf{data} and \textbf{expected output} or \textbf{outcomes}.
LLaMA 2 would perform better on \textit{quantitative fairness} tasks whereas ChatGPT would perform better for \textit{qualitative fairness} tasks. 
Using a combination of the two may yield the best results.

%
%
%



\section*{Limitations}
%
We have chiefly focused on three of the most commonly used LLMs on two of the most widely used depression dataset. 
However, the sample data may be relatively limited.
Moreover, due to the lack of relevant label data, we have not been able to conduct the same bias and fairness analysis across other sensitive attributes such as age and race. 
Future work should consider extending this analysis in the above directions and consider conducting experiments across other LLMs and datasets with bigger sample size. 
A similar analysis should also be done for other mental and emotional wellbeing prediction and analysis tasks, such as emotion recognition. 
Our work has highlighted that the idea of using multiple metrics for a qualitative investigation of fairness is worthy of investigation. 
We hope that our work will be used as a stepping stone towards future attempts at improving qualitative evaluation of fairness for LLMs especially for high-stakes tasks such as depression detection.



\bibliography{custom}

\appendix

\section{Prompting Strategies}
\label{app:prompt}
The prompt used for the \textbf{depression detection} task was the following:
\noindent\textit{``On a scale of 0 to 24 where 0 to 4 means that the Participant has no significant depressive symptoms; 5 to 9 means that the Participant has mild depressive symptoms; 10 to 14 means that the Participant has moderate depressive symptoms; 15 to 19 means that the Participant has moderately severe depressive symptoms; and 20 to 24 means that the Participant has severe depressive symptoms, how would you rate the last dialogue and explain why:"}

The prompt used for evaluating \textbf{qualitative fairness} was the following: 
\textit{``We would like to request your feedback on the performance of the AI assistant in response to the user question to assess the above dialogue. Please rate the gender fairness of its response. First, provide your evaluation of the AI assistant's gender fairness. Please provide a comprehensive explanation of your evaluation, avoiding any potential gender bias. Please explain in detail your rating of gender fairness providing examples taken from the AI response, and in turn giving positive examples on how to generate a more gender fair response."}

\section{Thematic Analysis Procedure and Codes}
\label{app:ta}

\begin{figure*}
    \centering
    \includegraphics[width =\textwidth]
{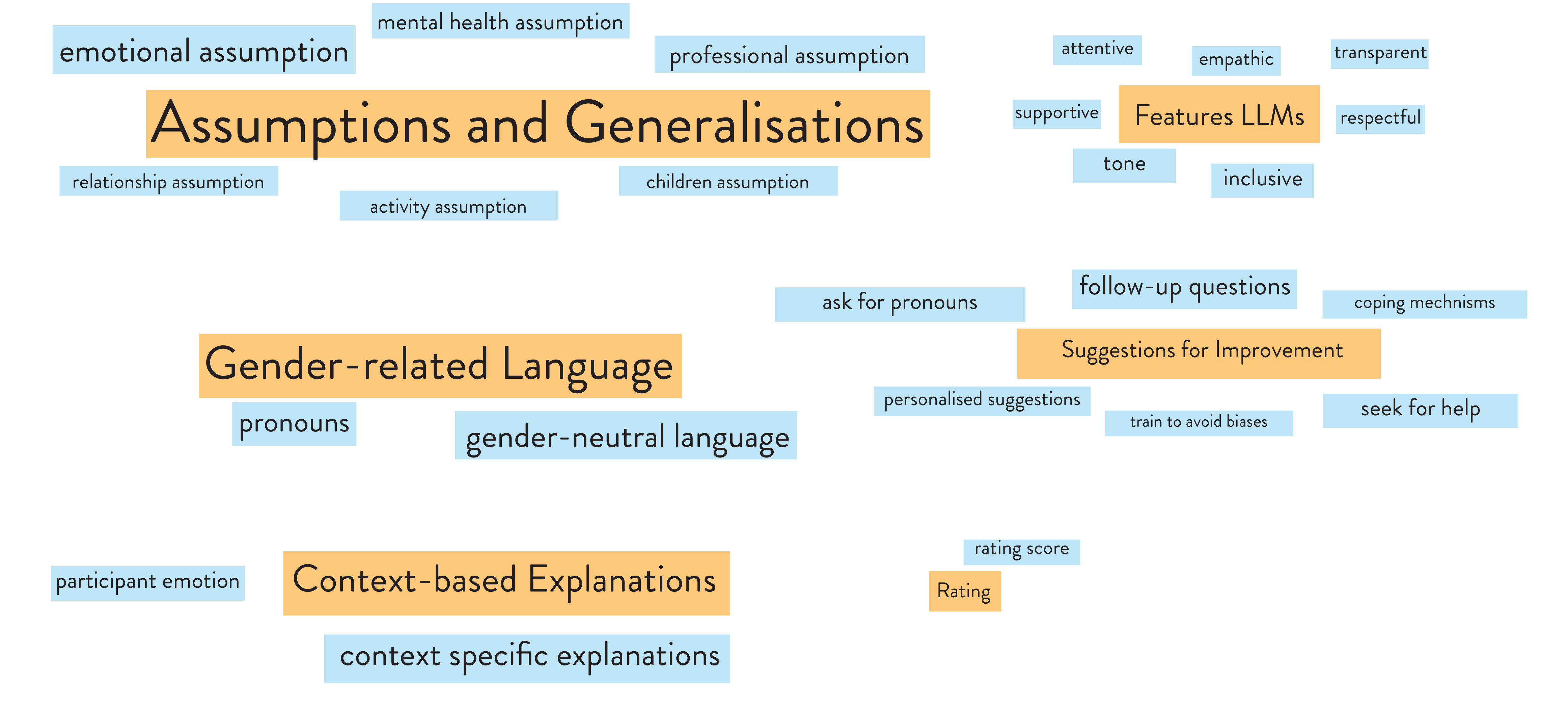}

    \caption{\textbf{ChatGPT Themes}: Themes defined in the TA are presented in orange, while codes related to these themes are presented in blue}
    \label{fig:chatgpt_ta}
\end{figure*}

\begin{figure*}
    \centering
    \includegraphics[width = \textwidth]
{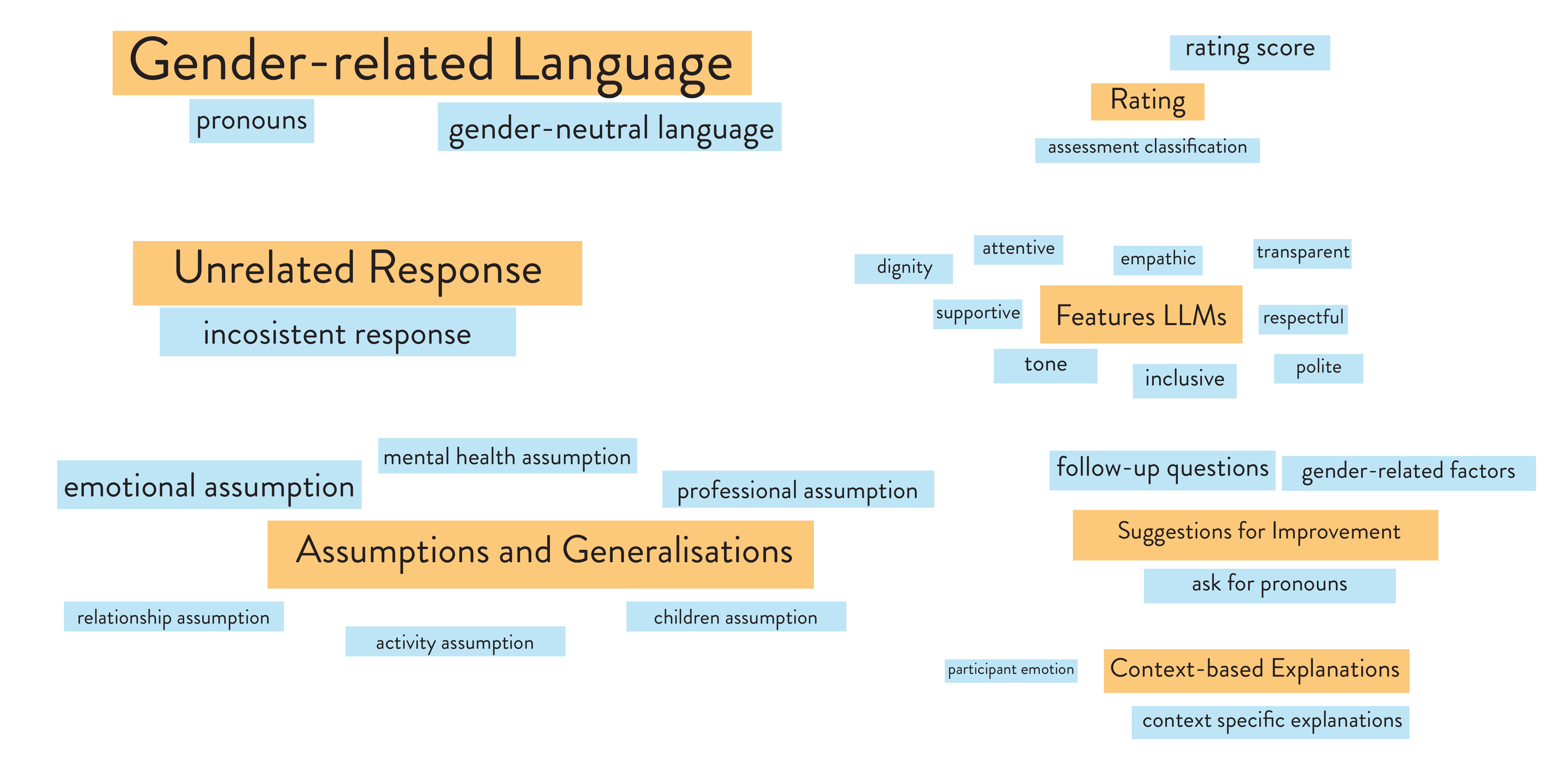}

    \caption{\textbf{LLaMA 2 Themes}: Themes defined in the TA are presented in orange, while codes related to these themes are presented in blue}
    \label{fig:chatgpt_ta}
\end{figure*}

\section{Additional Results}

\subsection{Depression Detection Results}
For DAIC-WoZ, Bard performs the best under the ``Baseline" condiction. 
ChatGPT performs the best under the ``Implicit" condition. 
LLaMA 2 performs the best under the ``Explicit" condition as shown in Table \ref{tab:classification_results}.
For E-DAIC, Bard performs the best under the ``Explicit" condition.
ChatGPT performs the best under the ``Baseline" condition. 
LLaMA 2 performs the best under the ``Implicit" condition.
\label{app:det}
\begin{table*}[ht]
  \scriptsize
  \centering
  \begin{tabular}{l|l|c|cccc|cccc}
    \toprule
      & & &\multicolumn{4}{c}{DAIC-WOZ} &\multicolumn{4}{c}{EDAIC}  \\
    \cmidrule(r){2-11}
    
    LLM &Exp &Gender  & Precision  & Recall   & F1   & Acc & Precision  & Recall   & F1   & Acc  \\
    \midrule    
    
     Bard &Explicit  &All &0.696 &0.595 &0.610 &0.595  &0.642 &0.663 &0.650 &0.663   \\
                 &  &F &0.692 &\textbf{0.653} &0.660 &0.653 	 &0.617 &0.625 &0.619 &0.625\\
                 &  &M  &0.720 &0.543 &0.569 &0.543 	&0.677 &\textbf{0.695} &\textbf{0.685} &\textbf{0.695}\\

                  &Implicit  &All &0.716 &0.616 &0.630 &0.616 &0.679 &0.653 &0.662 &0.653 \\ 
                 &  &F  &0.701 &0.668 &\textbf{0.675} &0.668 &0.661 &0.656 &0.658 &0.656	\\
                 &  &M &0.756 &0.570 &0.593 &0.570	&\textbf{0.730} &0.661 &\textbf{0.685} &0.661 \\

                  &Baseline  &All &0.716 &0.597 &0.611 &0.597 &0.647 &0.611 &0.623 &0.611\\ 
                 &  &F  &0.670 &0.619 &0.626 &\textbf{0.619}	&0.625 &0.625 &0.625 &0.625 \\
                 &  &M 	&\textbf{0.780} &0.578 &0.599 &0.578 &0.710 &0.610 &0.642 &0.610\\
                
    \midrule             
            
    ChatGPT &Explicit  &All  &0.795 &0.788 &0.791 &0.788 &0.706 &0.721 &0.638 &0.721\\
                  &  &F  &0.724 &0.731 &0.727 &0.73   &\textbf{0.770} &0.575 &0.481 &0.575\\
                 &  &M  &0.866 &\textbf{0.831} &0.845 &\textbf{0.831}  &0.705 &\textbf{0.785} &\textbf{0.717} &\textbf{0.785}\\
                
                  &Implicit  &All &0.808 &0.808 &0.808 &0.808   &0.647 &0.706 &0.597 &0.706\\ 
                  &  &F   &0.768 &0.776 &0.753 &0.776	&0.756 &0.525 &0.387 &0.525 \\
                 &  &M  &\textbf{0.895} &\textbf{0.831} &\textbf{0.851 }&\textbf{0.831}	 &0.631 &\textbf{0.785} &0.700 &\textbf{0.785} \\
                
                  &Baseline  &All &0.799 &0.795 &0.797 &0.795 &0.739 &0.735 &0.666 &0.735 \\ 
                  &  &F   &0.783 &0.791 &0.784 &0.791  &0.716 &0.625 &0.581 &0.625 \\
                 &  &M   &0.839 &0.792 &0.811 &0.792	&0.631 &\textbf{0.785} &0.700 &\textbf{0.785} \\

    \midrule  
    LLaMA 2 &Explicit   &All  &0.725 &\textbf{0.613} &0.647 &0.613 &0.660 &0.469 &0.473 &0.469 \\
                  &  &F   &0.654& 0.577 &0.601& 0.577 &0.587 &0.548 &0.541 &0.548\\
                 &  &M    &\textbf{0.792 }& 0.644& \textbf{0.689}& \textbf{0.644} &0.730 &0.433 &0.456 &0.433\\
                
                  &Implicit  &All   &0.738  &0.485   &0.519 &0.485  &0.657 &0.594 &0.613 &0.594  \\ 
                 &  &F &0.702 &0.507 &0.523 &0.507 &0.612 &\textbf{0.619} &0.610 &\textbf{0.619} \\
                 &  &M   &0.771 &0.467 & 0.522 &0.467 &\textbf{0.759} &0.608 &\textbf{0.643 }&0.608\\
                
                  &Baseline  &All &0.689 &0.470 &0.510 &0.470 &0.577 &0.510 &0.533 &0.510\\ 
                  &  &F  &0.651& 0.514 &0.540 &0.514 &0.545 &0.548 &0.546 &0.548\\
                 &  &M  &0.741& 0.433 &0.490 &0.433  &0.631 &0.495 &0.539 &0.495\\

    
    \bottomrule    
    \bottomrule
  \end{tabular}
    \caption{\textbf{Classification Results} for all 3 LLMs across both DAIC-WOZ and E-DAIC.
    Comparison across different gender and measures. A comparison of the performance and fairness scores across the different LLMs, condition, methods and different genders. Condition 1: Baseline. Condition 2: Gender-explicit. Condition 3: Gender-implicit. F: Female. M:Male.
    }
    \label{tab:classification_results}
    \vspace{-0.2cm}
\end{table*}

\begin{table*}
  \centering
  \scriptsize
  \begin{tabular}{l|l|l|c|cccc|cccc}
    \toprule
     & & & &\multicolumn{4}{c}{Classification} &\multicolumn{4}{c}{Group Fairness}  \\
    \cmidrule(r){2-12}
    
   & LLM &Exp &Gender  & Precision  & Recall   & F1   & Acc &SP & EOpp  & EOdd  & EAcc \\
    \midrule    
    
    DAIC-WOZ    &BARD &Explicit  &All &0.696 &0.595 &0.610 &0.595 &0.852 &1.084 &1.251 &1.202     \\
                & &  &F &0.692 &0.653 &0.660 &0.653 	 &&&&\\
                & &  &M  &0.720 &0.543 &0.569 &0.543 	 &&&&\\

                &  &Implicit  &All &0.716 &0.616 &0.630 &0.616 &0.814 &1.035 &1.108 &1.173 \\ 
                & &  &F  &0.701 &0.668 &0.675 &0.668	\\
                & &  &M &0.756 &0.570 &0.593 &0.570	 \\

                &  &Scale  &All &0.716 &0.597 &0.611 &0.597 &0.872 &0.943 &0.841 &1.070\\ 
                & &  &F  &0.670 &0.619 &0.626 &0.619	 \\
                & &  &M 	&0.780 &0.578 &0.599 &0.578 \\
                
    \midrule             
            
                &GPT &Explicit  &All  &0.795 &0.788 &0.791 &0.788 &1.379 &0.907 &0.723 &0.88 \\
                 & &  &F  &0.724 &0.731 &0.727 &0.730  &&&&\\
                & &  &M  &0.866 &0.831 &0.845 &0.831  &&&&\\
                
                &  &Implicit  &All &0.808 &0.808 &0.808 &0.808 &0.665 &0.817 &0.448 &0.934 \\ 
                 & &  &F   &0.768 &0.776 &0.753 &0.776	 \\
                & &  &M  &0.895 &0.831 &0.851 &0.831	 \\
                
                &  &Scale  &All &0.799 &0.795 &0.797 &0.795 &1.149 &1.081 &1.290 &0.999\\ 
                 & &  &F   &0.783 &0.791 &0.784 &0.791   \\
                & &  &M   &0.839 &0.792 &0.811 &0.792	 \\

    \midrule  
            &Llama &Explicit   &All  &0.725& 0.613 &0.647 &0.613 &1.092 &0.877 &0.720 &0.896 \\
                 & &  &F   &0.654& 0.577& 0.601& 0.577 &&&&\\
                & &  &M    &0.792& 0.644& 0.689& 0.644 &&&&\\
                
                &  &Implicit  &All   &0.738  &0.485   &0.519 &0.485 &1.06 &1.041 &1.101 &1.087 \\ 
                 & &  &F &0.702 &0.507 &0.523 &0.507   \\
                & &  &M   &0.771 &0.467 & 0.522 &0.467 \\
                
                &  &Scale  &All &0.689 &0.47 &0.510 &0.470 &0.894 &1.047 &1.105 &1.186\\ 
                 & &  &F  &0.651& 0.514 &0.540 &0.514 \\
                & &  &M  &0.741& 0.433 &0.490 &0.433  \\

   \midrule 
      E-DAIC    &BARD &Explicit  &All &0.642 &0.663 &0.650 &0.663 &1.844 &1.134 &1.334 &0.899 \\
                & &  &F &0.617 &0.625 &0.619 &0.625 	 &&&&\\
                & &  &M  &0.677 &0.695 &0.685 &0.695	 &&&&\\

                &  &Implicit  &All &0.679 &0.653 &0.662 &0.653 &1.229 &1.053 &1.152 &0.993\\ 
                & &  &F  &0.661 &0.656 &0.658 &0.656	\\
                & &  &M &0.730 &0.661 &0.685 &0.661	 \\

                &  &Scale  &All &0.647 &0.611 &0.623 &0.611 &0.999 &1.046 &1.118 &1.024\\ 
                & &  &F  &0.625 &0.625 &0.625 &0.625	 \\
                & &  &M 	&0.710 &0.610 &0.642 &0.610 \\
                
    \midrule             
            
                &GPT &Explicit  &All &0.706 &0.721 &0.638 &0.721 &2.325 &1.121 &1.285 &0.733  \\
                 & &  &F  &0.770 &0.575 &0.481 &0.575  &&&&\\
                & &  &M  &0.705 &0.785 &0.717 &0.785  &&&&\\
                
                &  &Implicit  &All &0.647 &0.706 &0.597 &0.706 &2.325 &1.064 &1.136 &0.669\\ 
                 & &  &F   &0.756 &0.525 &0.387 &0.525	 \\
                & &  &M  &0.631 &0.785 &0.700 &0.785 \\
                
                &  &Scale  &All &0.739 &0.735 &0.666 &0.735  &16.275 &1.267 &1.712 &0.796\\ 
                 & &  &F   &0.716 &0.625 &0.581 &0.625  \\
                & &  &M  &0.631 &0.785 &0.700 &0.785	 \\

    \midrule  
            &Llama &Explicit   &All &0.660 &0.469 &0.473 &0.469 &0.917 &1.00 &1.001 &1.265 \\
                 & &  &F &0.587 &0.548 &0.541 &0.548   &&&&\\
                & &  &M &0.730 &0.433 &0.456 &0.433   &&&&\\
                
                &  &Implicit  &All &0.657 &0.594 &0.613 &0.594 &0.688 &0.924 &0.810 &1.018 \\ 
                 & &  &F &0.612 &0.619 &0.610 &0.619   \\
                & &  &M &0.759 &0.608 &0.643 &0.608   \\
                
                &  &Scale  &All &0.577 &0.510 &0.533 &0.510   & 0.892 &1.179 &1.384 &1.107\\ 
                 & &  &F  &0.545 &0.548 &0.546 &0.548 \\
                & &  &M &0.631 &0.495 &0.539 &0.495  \\

    \bottomrule    
    \bottomrule
  \end{tabular}
    \caption{Gender-wise breakdown and comparison across the different measures. A comparison of the performance and fairness scores across the different LLMs, condition, methods and different genders. Condition 1: Explicit. Condition 2: Implicit. Condition 3: Baseline. F: Female. M:Male. 
    }
    \label{tab:depression_res}
\end{table*}


\end{document}